\documentclass{article}

\usepackage[utf8]{inputenc}
\usepackage[english]{babel}
\usepackage{bm}
\usepackage{microtype}
\usepackage{changepage}

\usepackage{amsmath,amsfonts,mathtools,amsthm}
\usepackage{mathptmx}       
\usepackage{helvet}         
\usepackage{courier}        
\newtheorem{theorem}{Theorem}
\usepackage{graphicx}        
\usepackage{booktabs}
														
\usepackage{tikz}
\usetikzlibrary{plotmarks}
\usepackage{pgfplots}
\usepackage{pgfplotstable}
\pgfplotsset{compat=1.12}

\usetikzlibrary{shapes,arrows}
\usetikzlibrary{arrows}
\usetikzlibrary{decorations}

\usetikzlibrary{decorations.pathmorphing}

\usepackage{xspace}
\usepackage[algo2e,ruled,vlined,linesnumbered]{algorithm2e}
\usepackage[numbers,sort&compress]{natbib}
\usepackage{pdflscape}
\usepackage[affil-it]{authblk}
\usepackage{titling}
\usepackage{footnote}
\usepackage[shortcuts]{extdash}

\makesavenoteenv{tabular}
\makesavenoteenv{table*}

\usepackage{hyperref}

\newcommand{\needle}{\textsc{Needle}\xspace}
\newcommand{\eg}{e.\,g.\xspace}
\newcommand{\ie}{i.\,e.\xspace}
\newcommand{\wrt}{w.\,r.\,t.\xspace}

\allowdisplaybreaks[4]
\newcommand{\cga}{cGA\xspace}
\newcommand{\umda}{UMDA\xspace}
\newcommand{\pbil}{PBIL\xspace}
\newcommand{\fda}{FDA\xspace}
\newcommand{\mmasib}{MMAS$_\text{ib}$\xspace}

\newcommand{\oea}{(1+1)~EA\xspace}

\newcommand{\om}{\textsc{OneMax}\xspace}
\newcommand{\onemax}{\om}
\newcommand{\lo}{\textsc{LeadingOnes}\xspace}

\newcommand{\bv}{\textsc{BinVal}\xspace}
\newcommand*{\sigcGA}{sig\=/cGA\xspace}

\newcommand{\R}{\ensuremath{\mathbb{R}}}

\newcommand{\N}{\ensuremath{\mathbb{N}}}

\DeclareMathOperator{\poly}{poly}

\newcommand*{\nblEDA}{$n$-Bernoulli-$\lambda$-EDA\xspace}

\makeatletter
\def\NAT@spacechar{~}
\makeatother

\begin{document}

\title{Theory of Estimation-of-Distribution Algorithms}
\author{Martin S. Krejca\thanks{Hasso Plattner Institute, University of Potsdam, Potsdam, Germany\\ \hspace*{0.8 em}\emph{e-mail: \href{mailto:martin.krejca@hpi.de}{martin.krejca@hpi.de}}}\hspace*{0.5 em} and Carsten Witt\thanks{DTU Compute, Technical University of Denmark, Kgs.~Lyngby, Denmark\\ \hspace*{0.8 em}\emph{e-mail: \href{mailto:cawi@dtu.dk}{cawi@dtu.dk}}}}
\date{\vspace*{-1 cm}}
%
%
\maketitle

\begin{abstract}
Estimation-of-distribution algorithms (EDAs) are general metaheuristics used in optimization that represent 
a more recent alternative to classical approaches like evolutionary algorithms. In a nutshell, EDAs 
typically do not directly evolve populations of search points but 
build probabilistic models 
of promising solutions by repeatedly sampling and selecting  points from the underlying search space. Recently, 
there has been made significant progress in the theoretical understanding of EDAs.
This article provides an up-to-date overview of the  most commonly analyzed EDAs 
and 
the most recent theoretical results in this area. In particular, emphasis is put on 
the runtime analysis of simple univariate EDAs, including a description of typical benchmark 
functions and tools for the analysis. 
Along the way, open problems and 
directions for future research are described.
\end{abstract}

\section{Introduction}
\label{sec:eda:introduction}

Optimization is one of the most important fields in computer science, with many problems being NP-hard and thus not necessarily easy to solve. Hence, \emph{heuristics} play a major role, i.e., optimization algorithms that try to yield solutions of good quality in a reasonable amount of time. Research over the past decades has resulted in many good heuristics developed for classical NP-hard problems. Unfortunately, these heuristics are tailored with specific problems in mind and exploit certain problem-specific properties in order to save computation time. Thus, they cannot be used for problems that do not feature these specific properties.

One alternative to problem-specific heuristics are \emph{general-purpose} heuristics. The information about the problem to optimize that these algorithms have access to is fairly limited, up to the point that they are only able to compare the quality of different solutions relatively.
This has the advantage that the problem itself does not have to be formalized but only the quality of a solution, as the problem formalization is communicated implicitly via the quality measure to the algorithm. In turn, this results in great reusability of the algorithm for different problems. 

One class of general-purpose heuristics are \emph{evolutionary algorithms} (EAs)~\cite{CompIntHandbook2015}.
EAs are characterized by creating new solutions from already generated solutions. Oftentimes, many solutions are stored and only changed \emph{(evolved)} locally, preferably discarding bad solutions and saving good ones. Such algorithms are EAs in the classical sense~\cite{Simon2013}.

The concept of EAs can be broadened when being less restrictive about what is being evolved. A similar approach to changing solutions directly is to instead change the procedure that generates the solutions in the first place. Thus, a solution-generating mechanism is evolved. Algorithms following this approach are called \emph{estimation-of-distribution algorithms} (EDAs)~\cite{LarranagaLozanoEDABook, PelikanSastryCantuPaz2006, HauschildPelikan11, PelikanHandbook15}. They are not EAs in the classical sense but can be considered EAs in the broad sense, as just described.

EDAs have been used very successfully in real-world applications~\cite{LarranagaLozanoEDABook, PelikanSastryCantuPaz2006, HauschildPelikan11, PelikanHandbook15} and recently gathered momentum in the theory community analyzing EAs~\cite{DangLehreGECCO15, DBLP:conf/isaac/FriedrichKKS15, FriedrichKK16EDA, SudholtWittGECCO2016, KrejcaWittFOGA2017, LehreNguyenGECCO17, WittGECCO17}. The aim of theoretically analyzing EAs is to provide guarantees for the algorithms and to gain insights into their behavior in order to optimize the algorithms themselves. Common guarantees include the expected time until an algorithm finds a solution of sufficient quality, the probability to do so after a certain time, or the fact that the algorithm is even able to find desired solutions.

With this article, we provide a state-of-the-art overview about the theoretical results on EDAs for \emph{discrete} domains, as that is their main field of application. To the best of our knowledge, while continuous EDAs exist, no detailed theoretical analyses have been conducted so far. We present the most commonly investigated EDAs and give an outline of the history of their analyses, providing deep insights into some of the latest results. After reading, the reader should be familiar with EDAs in general, the current state of theoretical research, and common tools used for the analyses.

In Section~\ref{sec:eda:def-eda}, we go more into detail about how EDAs work, we introduce the scenario used in most theoretical papers, and we provide different ways of classifying EDAs, stating the most commonly analyzed algorithms. Further, we mention some tools that are often used when deriving results for EDAs. Then, in Section~\ref{sec:eda:fitness_functions}, we give a short overview over the most commonly considered objective functions. In Section~\ref{sec:eda:convergence}, we discuss the historically older results of convergence analyses on EDAs. After that, in Section~\ref{sec:eda:runtime}, we present more recent results on EDAs, which consider the actual runtime of an algorithm. We end this article in Section~\ref{sec:eda:conclusions} with some conclusions and open problems.

\section{Estimation-of-Distribution Algorithms}
\label{sec:eda:def-eda}

In general, EDAs are problem-agnostic optimization algorithms that store a probabilistic model over the solution space. This model is the core part of these algorithms. It implies a probability distribution over the solution space and is iteratively refined, using samples. Ideally, the model converges to a state that only produces optimal solutions.

Since EDAs make use of sample sets~-- so-called \emph{populations}~--, they are quite similar in this respect to EAs. However, the main difference is that EAs exclusively store a population and progress using solely this information: by varying samples~-- so-called \emph{individuals}~-- from the population. Thus, they have quite a local view over the solution space and advance locally. In contrast, the probabilistic model of an EDA models most of the time the entire solution space. Updates to the model are done using the old model as well as a population. Hence, EDAs employ a more general view over the solution space than classical EAs.

The probabilistic model of an EDA is used as an \emph{implicit} probability distribution over the solution space, instead of an explicit distribution. This is usually done by constraining the distributions that can be modeled and by factorizing them, i.e., by writing the distribution as a product of marginal probabilities. \citet{HauschildPelikan11} distinguish between many different classes of EDAs with respect to how strongly constrained the models are. An advantage of factorizing a distribution is to save a lot of memory, since an explicit distribution would necessitate to store a probability for each solution, which is not feasible. With a factorization, only the factors have to be stored in memory. However, even then it is possible for the model to grow to sizes exponential in the input~\cite{GaoCulbersonSpaceEDAs05}.

As mentioned above, EDAs also use populations, like EAs, sampled from their probabilistic model, in order to update said model. It is up to the EDA to decide what to do with its population. However, all so far theoretically analyzed EDAs have in common that they always discard their population after every iteration, valuing the model higher than the population.

In the following, we first state the optimization domain for EDAs that we consider in this chapter. Then we discuss different classifications of EDAs and name various algorithms that fall into those classes. Last, we mention the tools that are commonly used in the current theoretical research of EDAs.

\subsection{Scenario}
As in the theory of EAs, theoretical analyses of EDAs mainly consider pseudo-Boolean optimization, i.e., optimization of a function $f\colon \{0, 1\}^n \to \R$, often referred to as \emph{fitness function}. Conventionally, the function value of a bit string~$\bm{x}$ is called the \emph{fitness of~$\bm{x}$}.

The aspect of an EDA being a general-purpose solver is modeled as a classical black-box setting, where the algorithm only gains problem-specific information from querying the fitness function by inputting bit strings and receiving their respective fitness. In this setting, mostly two different scenarios have been of major interest:
\begin{description}
    \item[\textbf{Convergence Analyses}] In this historically older topic, EDAs have been analyzed with respect to the convergence of their probabilistic model, i.e., if they succeed at all in optimizing certain fitness functions. We discuss this scenario in more depth in Section~\ref{sec:eda:convergence}.
    
    \item[\textbf{Runtime Analyses}] A more recent trend is the analysis of an EDA’s runtime on certain functions. In this scenario, the focus is the number of queries needed until an optimum or a solution of sufficient quality is sampled, i.e., the first hitting time of an algorithm sampling such a solution. Although sampling a desired solution can happen by chance, the analyses usually entail that the probabilistic model of an EDA makes it very likely for said solution to be sampled again. Section~\ref{sec:eda:runtime} goes into detail about this topic.
\end{description}

\subsection{Classifications of EDAs}
Arguably, the most straightforward way of classifying EDAs is with respect to the power of their underlying probabilistic model: \emph{univariate} algorithms only use a single variable in their model per problem variable\footnote{In our setting of pseudo-Boolean optimization, a \emph{problem variable} is a position in a bit string, i.e., one dimension of the hypercube.}\!. In contrast, \emph{multivariate} algorithms use more than a single variable to model a problem variable. Thus, univariate EDAs are not able to capture dependencies between problem variables, whereas multivariate EDAs are explicitly constructed to do so.

\citet{PelikanHandbook15} give a more fine-grained classification of EDAs, differentiating multivariate EDAs even further with respect to how many dependencies can be captured by the underlying probabilistic model.


Note that the classification into univariate and multivariate EDAs does not constrain the populations at all.

\subsubsection{Univariate Algorithms}
\label{sec:eda:univariate}
When optimizing a pseudo-Boolean function, univariate EDAs assume independence of all of the~$n$ different bit positions to optimize. Under this assumption, every probability distribution can be factorized into a product of~$n$ different probabilities~$\bm{p}_i$, collected in a vector~$\bm{p}$ of length~$n$. A bit string~$\bm{x}$ is then sampled by choosing each bit~$\bm{x}_i$ to be~$1$ with probability~$\bm{p}_i$ and~$0$ otherwise. Since each~$\bm{p}_i$ determines how frequently, in expectation, a~$1$ is sampled at position~$i$, we call these probabilities \emph{frequencies}, following the common naming convention (\cite{FriedrichKK16EDA}). The vector~$\bm{p}$ is then consequently called \emph{frequency vector}.

\paragraph{\nblEDA}{
Although the class of univariate EDAs does not limit the populations of the algorithms in any way, the most commonly considered univariate EDAs discard their entire population after every iteration. Thus, from a theoretical point of view, a run of a univariate EDA can be modeled as a series $(\bm{p}^{(t)})_{t \in \N}$ of frequency vectors over the number of iterations~$t$. Usually, $\bm{p}^{(0)}$ models the uniform distribution by satisfying that $\bm{p}^{(0)}_i = 1/2$ for each~$i$. \citet{FriedrichKK16EDA} capture this class of univariate EDAs in a framework called the \emph{\nblEDA} (Algorithm~\ref{alg:eda:nblEDA}).

The \nblEDA samples~$\lambda$ individuals each iteration and performs an update to its frequency vector, using the current frequency vector as well as all of the just sampled individuals and their respective fitness. The function performing this update is called the \emph{update scheme} and fully characterizes the algorithm.

Note that we do not specify a termination criterion. In fact, determining what a good criterion is may vary between different use cases of the algorithm. When considering the expected runtime of these algorithms (Section~\ref{sec:eda:runtime}), we are interested in the number of fitness function evaluations until an optimal solution is sampled for the first time.

\begin{algorithm2e}
    \caption{\nblEDA with a given update scheme $\varphi$, optimizing~$f$}
    \label{alg:eda:nblEDA}
    $t \gets 0$\;
    \lForEach{$i \in \{1, \ldots, n\}$}
    {
        $\bm{p}^{(t)}_i \gets \frac{1}{2}$
    }
    \Repeat{\emph{termination criterion met}}
    {
        $D \gets \emptyset$\;
        \ForEach{$j \in \{1, \ldots, \lambda\}$}
        {
            $\bm{x} \gets$ offspring sampled with respect to $\bm{p}^{(t)}$\;
            $D \gets D \cup \{\bm{x}\}$;\,\footnotemark
        }
        $\bm{p}^{(t + 1)} \gets \varphi\big(\bm{p}^{(t)}, \big(\bm{x}, f(\bm{x})\big)_{\bm{x} \in D}\big)$\;\label{line:eda:update}
        $t \gets t + 1$\;
    }
\end{algorithm2e}
\footnotetext{Note that~$D$ is a multiset, that is, we allow duplicates.}

In many EDAs, if a frequency is either~$0$ or~$1$, all bits sampled at the respective position will be~$0$ or~$1$, respectively, and the update scheme will not change the frequency anymore. To prevent this, the algorithm is usually modified such that each frequency is only allowed to take values in an interval $[m, 1 - m] \subset [0, 1]$, where $m \in (0, 1/2]$ is called a \emph{margin}; the values~$m$ and $1 - m$ are called \emph{borders}. Usually, a margin of~$1/n$ is chosen~\cite{ChenEtAlCEC09b, Neumann2010a, DangLehreGECCO15}. In a scenario with a margin, Line~\ref{line:eda:update} of Algorithm~\ref{alg:eda:nblEDA} can be modified as follows:
\[
    \textbf{foreach } i \in \{1, \ldots, n\} \textbf{ do } \bm{p}^{(t + 1)}_i \gets \max\!\Big\{\!m,\ \min\!\big\{\!1 - m,\ \varphi\big(\bm{p}^{(t)}, \big(\bm{x}, f(\bm{x})\big)_{\bm{x} \in D}\big)_i\big\}\Big\};
\]

We continue to give an overview of the most commonly theoretically analyzed univariate EDAs and show how they fit into the \nblEDA framework. We present the algorithms without a margin although they are commonly analyzed with a margin of~$1/n$.
}

Since many of the following examples do not make use of the entire population of size~$\lambda$ (the \emph{population size}) but select a certain number~$\mu$ (the \emph{effective population size}) of individuals according to their fitness values, we denote the $k$th-best individual as $\bm{x}^{(k)}$\!, where $1 \leq k \leq \mu$; ties are broken \emph{uniformly at random}. Thus, $\bm{x}^{(1)}$ denotes an individual with best fitness.

\paragraph{\umda}{
The arguably easiest update scheme is given by the \emph{Univariate Marginal Distribution Algorithm} (\umda; Algorithm~\ref{alg:eda:umda})~\cite{MuhlenbeinPaass1996}. It samples~$\lambda$ individuals each iteration, of which~$\mu$ best are chosen. Then, each frequency~$\bm{p}_i$ is set to the relative frequency of~$1$s at position~$i$ in the set of the~$\mu$ best individuals, regardless of the current frequency.
\begin{algorithm2e}
    \caption{\umda with population size~$\lambda$, effective population size~$\mu$, optimizing~$f$}
    \label{alg:eda:umda}
    $t \gets 0$\;
    \lForEach{$i \in \{1, \ldots, n\}$}
    {
        $\bm{p}^{(t)}_i \gets \frac{1}{2}$
    }
    \Repeat{\emph{termination criterion met}}
    {
        $D \gets \emptyset$\;
        \ForEach{$j \in \{1, \ldots, \lambda\}$}
        {
            $\bm{x} \gets$ offspring sampled with respect to $\bm{p}^{(t)}$\;
            $D \gets D \cup \{\bm{x}\}$\;
        }
        \lForEach{$i \in \{1, \ldots, n\}$}
        {
            $\bm{p}^{(t + 1)}_i \gets \frac{1}{\mu}\sum_{k = 1}^{\mu}\bm{x}^{(k)}_i$\!
        }
        $t \gets t + 1$\;
    }
\end{algorithm2e}

The update scheme of \umda allows it to go from any valid frequency to any other in a single step if not stuck. Thus, the difference of two consecutive frequencies~$\bm{p}^{(t)}_i$ and~$\bm{p}^{(t + 1)}_i$ can only be trivially bounded by roughly~$1$. We call such a difference the \emph{step size} of the algorithm.
}

\paragraph{\pbil}{
A variant of \umda that has an adjustable step size is the \emph{Population-Based Incremental Learning} algorithm (\pbil; Algorithm~\ref{alg:eda:pbil})~\cite{Baluja1994}. A frequency is updated in a way similar to \umda, but the new frequency is a convex combination with parameter~$\rho$ of the current frequency and the relative frequencies of~$1$s at that position. Thus, the step size is now bounded by~$\rho$, and \umda is a special case of \pbil with $\rho = 1$.

\begin{algorithm2e}
    \caption{\pbil with population size~$\lambda$, effective population size~$\mu$, and learning rate~$\rho$, optimizing~$f$}
    \label{alg:eda:pbil}
    $t \gets 0$\;
    \lForEach{$i \in \{1, \ldots, n\}$}
    {
        $\bm{p}^{(t)}_i \gets \frac{1}{2}$
    }
    \Repeat{\emph{termination criterion met}}
    {
        $D \gets \emptyset$\;
        \ForEach{$j \in \{1, \ldots, \lambda\}$}
        {
            $\bm{x} \gets$ offspring sampled with respect to $\bm{p}^{(t)}$\;
            $D \gets D \cup \{\bm{x}\}$\;
        }
        \lForEach{$i \in \{1, \ldots, n\}$}
        {
            $\bm{p}^{(t + 1)}_i \gets (1 - \rho)\bm{p}^{(t)}_i + \frac{\rho}{\mu}\sum_{k = 1}^{\mu}\bm{x}^{(k)}_i$\!
        }
        $t \gets t + 1$\;
    }
\end{algorithm2e}
}

\paragraph{\mmasib}{
Another important univariate EDA is \emph{Max-Min Ant System with iteration-best update} (\mmasib; Algorithm~\ref{alg:eda:mmas_ib})~\cite{Neumann2010a}, which is a special case of \pbil when setting $\mu = 1$, i.e., when we only consider the best individual in each iteration. \mmasib also falls into the general class of \emph{Ant Colony Optimization} (ACO) algorithms~\cite{DorManCol1991:TR-POLI-91-016}. 
Although, ACO spans an entire research topic independent of EDAs and is typically not considered to be an EDA, the process of how it produces solutions iteratively can be viewed as refining a probabilistic model. Thus, we view ACO as an EDA.

ACO considers graphs whose edges are weighted with probabilities: so-called \emph{pheromones}. Additionally, the algorithm uses agents~-- so-called \emph{ants}~-- that traverse the graph and thus construct paths. At each vertex~$v$, if a path should be extended, an ant chooses an edge with a certain probability with respect to the pheromones on all of the outgoing edges of~$v$. After the data of all ants has been collected, all pheromones decrease (they \emph{evaporate}) and some are increased afterward, usually the ones that are part of the best solutions constructed.

When considering pseudo-Boolean optimization, a graph for ACO can be modeled as a multigraph with~$n + 1$ vertices from~$0$ to~$n$, each vertex having exactly two outgoing edges to its direct successor (except for vertex~$n$; see Figure~\ref{fig:eda:construction-multigraph}). One of these edges is interpreted as a~$0$, the other one as a~$1$. Each solution is constructed by letting an ant traverse the graph starting at~$0$ and ending at~$n$. The corresponding edges are then interpreted as a bit string of length~$n$. Note how the probability to choose an edge corresponding to a~$1$ is equal to an \nblEDA's frequency for that respective position.

\begin{figure*}
    \begin{center}
        \begin{tikzpicture}[scale=2]
        \tikzstyle{antbody}=[fill=black,draw=black];
        \tikzstyle{antleg}=[black,rounded corners=0pt];
        \begin{scope}[shift={(-0.55,0)},scale=0.2]
        \draw[antleg] (6pt+5pt,0pt) -- ++(-5pt,+7pt) -- ++(-11pt,+1pt) -- ++(-3pt,+1pt);
        \draw[antleg] (6pt+8pt,0pt) -- ++(+1pt,+7pt) -- ++(-3pt,+7pt);
        \draw[antleg] (6pt+10pt,0pt) -- ++(+8pt,+6pt) -- ++(+2pt,+5pt);
        \draw[antleg,rounded corners=1pt] (6pt-2pt+2*9pt-1pt+6pt+4pt,0pt) -- ++(-2pt,+8pt) -- ++(+10pt,+3pt) -- ++(+4pt,+2pt);
        \draw[antleg] (6pt+5pt,0pt) -- ++(-5pt,-7pt) -- ++(-11pt,-1pt) -- ++(-3pt,-1pt);
        \draw[antleg] (6pt+8pt,0pt) -- ++(+1pt,-7pt) -- ++(-3pt,-7pt);
        \draw[antleg] (6pt+10pt,0pt) -- ++(+8pt,-6pt) -- ++(+2pt,-5pt);
        \draw[antleg,rounded corners=1pt] (6pt-2pt+2*9pt-1pt+6pt+4pt,0pt) -- ++(-2pt,-8pt) -- ++(+10pt,-3pt) -- ++(+4pt,-2pt);
        \filldraw[antbody] (6pt-2pt+9pt,0) ellipse (9pt and 2pt);
        \filldraw[antbody] (0,0) ellipse (7pt and 5pt);
        \filldraw[antbody] (6pt-2pt+2*9pt-1pt+6pt,0) ellipse (5pt and 4pt);
        \end{scope}
        \tikzstyle{node}=[black,fill=white,thick];
        \tikzstyle{edge}=[black,thick,-triangle 60];
        \foreach \x in {1,2,3,4,5} {
            \draw[edge,shorten >=8pt,shorten <=8pt] (\x-1,0) .. controls (\x-1+0.3,0.4) and (\x-1+0.7,0.4) .. (\x,0) node[pos=0.5,above] {$e_{\x,1}$};
            \draw[edge,shorten >=8pt,shorten <=8pt] (\x-1,0) .. controls (\x-1+0.3,-0.4) and (\x-1+0.7,-0.4) .. (\x,0) node[pos=0.5,below] {$e_{\x,0}$};
        }
        \foreach \x in {0,1,2,3,4,5} {
            \filldraw[node] (\x,0) circle (4pt) node {$v_{\x}$};
        }
        \end{tikzpicture}
    \end{center}
    \caption{The ACO graph for pseudo-Boolean optimization with $n=5$ bits.}
    \label{fig:eda:construction-multigraph}
\end{figure*}
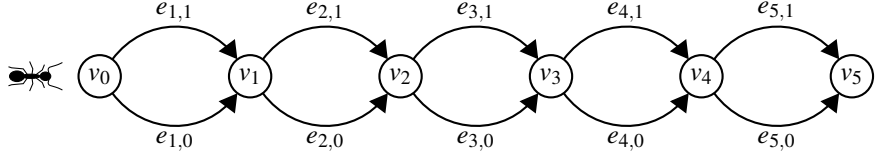

\mmasib is a variant of the Max-Min Ant System algorithm~\cite{StuetzleHoosMaxMin} that only makes an update with respect to the path of the best ant of each iteration, using a classical update rule in ACO.

\begin{algorithm2e}[h]
    \caption{\mmasib with population size~$\lambda$ and evaporation factor~$\rho$, optimizing~$f$}
    \label{alg:eda:mmas_ib}
    $t \gets 0$\;
    \lForEach{$i \in \{1, \ldots, n\}$}
    {
        $\bm{p}^{(t)}_i \gets \frac{1}{2}$
    }
    \Repeat{\emph{termination criterion met}}
    {
        $D \gets \emptyset$\;
        \ForEach{$j \in \{1, \ldots, \lambda\}$}
        {
            $\bm{x} \gets$ offspring sampled with respect to $\bm{p}^{(t)}$\;
            $D \gets D \cup \{\bm{x}\}$\;
        }
        \lForEach{$i \in \{1, \ldots, n\}$}
        {
            $\bm{p}^{(t + 1)}_i \gets (1 - \rho)\bm{p}^{(t)}_i + \rho\bm{x}^{(1)}_i$\!
        }
        $t \gets t + 1$\;
    }
\end{algorithm2e}
}

\paragraph{\cga}{
An algorithm with a different approach is the \emph{Compact Genetic Algorithm} (\cga; Algorithm~\ref{alg:eda:cga})~\cite{HarikLG98}. It samples exactly two individuals each iteration and compares their bit values component-wisely. If the bits at position~$i$ are the same, the frequency~$\bm{p}_i$ is left unchanged. Otherwise, the frequency is adjusted by $\pm 1/K$, where~$K$ is an algorithm-specific parameter, often referred to as \emph{population size}, such that the probability of sampling the bit value of the fitter individual is more likely in the next iteration.

\begin{algorithm2e}
    \caption{\cga with population size~$K$, optimizing~$f$}
    \label{alg:eda:cga}
    $t \gets 0$\;
    \lForEach{$i \in \{1, \ldots, n\}$}
    {
        $\bm{p}^{(t)}_i \gets \frac{1}{2}$
    }
    \Repeat{\emph{termination criterion met}}
    {
        $D \gets \emptyset$\;
        \ForEach{$j \in \{1, \ldots, \lambda\}$}
        {
            $\bm{x} \gets$ offspring sampled with respect to $\bm{p}^{(t)}$\;
            $D \gets D \cup \{\bm{x}\}$\;
        }
        \lForEach{$i \in \{1, \ldots, n\}$}
        {
            $\bm{p}^{(t + 1)}_i \gets \max\!\Big\{\!0,\ \min\!\big\{1,\ \bm{p}^{(t)}_i  + \frac{1}{K}\big(\bm{x}^{(1)} - \bm{x}^{(2)}\big)\big\}\Big\}$\!
        }
        $t \gets t + 1$\;
    }
\end{algorithm2e}
}


\subsubsection{Multivariate Algorithms}
The class of multivariate EDAs consists of all algorithms that can use multiple variables to model one problem variable and thus express dependencies. A compact representation of such dependencies can be modeled as a directed graph whose vertices are the variables and whose edges denote dependencies among the variables. For each vertex, the probability distribution conditional on all its adjacent vertices with an incoming edge (its \emph{parents}) is stored. This results in a factorization of the problem space that respects the given dependencies. Multivariate EDAs can assume a certain dependency model and only learn the respective (conditional) probabilities of the factorization, or they additionally try to learn a model that fits well to the samples.

The \emph{Factorized Distribution Algorithm} (FDA)~\cite{MuehlenbeinMahnigFDA-ECJ99} falls into the former category. It assumes a factorization according to a so-called \emph{additively decomposable function} (ADF), i.e., a function that is a sum of multivariate subfunctions. For each set of variables per subfunction, \fda creates a \emph{metavariable} and expresses the objective function (the ADF) with respect to those metavariables. In each iteration, it samples solutions with respect to the factorization, selects a subset of them, and estimates the conditional probabilities based on these samples.
Note that \fda is a generalization of the update of \umda and coincides with it if no dependencies between the problem variables exist.

Another approach also using metavariables is the \emph{Extended Compact Genetic Algorithm} (ECGA)~\cite{HarikLoboSastryECGA2006}. Different from \fda, a metavariable of ECGA represents multiple variables at once (i.e., it is assumed that such variables are strongly correlated). Each iteration, the algorithm starts by placing each problem variable into its own class. Then, it greedily merges two classes such that a certain metric (the so-called \emph{Bayesian information criterion}) is maximized, using samples from the current model. If no further improvement can be made, the merging process stops and the algorithm uses the newly created model.


The easiest of the multivariate cases is the one where each variable can be at most dependent on one other variable, i.e., a bivariate setting, and the arguably easiest probabilistic model in such a setting is a path. This model is used in the \emph{Mutual-Information-Maximization Input Clustering} (MIMIC) algorithm introduced by~\citet{BonetEtAlMIMIC1997}. The idea of the underlying model is to construct a path that minimizes the Kullback-Leibler divergence with respect to the bivariate setting, i.e., to find a permutation that can explain the sample data best. However, since there are $n!$ possible permutations for~$n$ variables, the authors suggest a greedy approach that makes use of the empirical entropies, i.e., the entropies of the sample data. First, a variable with minimal entropy is chosen as the start vertex of the path. Then, the path is continued by choosing a node that has minimal conditional entropy with respect to the currently last vertex in the path.

The \emph{Bivariate Marginal Distribution Algorithm} (BMDA)~\cite{Pelikan1999Bivariate} uses a somewhat similar approach. However, it does not consider paths as its model for dependency graphs but rather a forest of rooted trees. In order to determine which variables are dependent on which other variables, the Pearson's chi-square statistics is used as an indicator. If the indicator is too low, the corresponding variables are considered independent. The forest is then created greedily very similarly to regular algorithms for maximum spanning trees: iteratively, a vertex is added to one of the trees that has maximum Pearson chi-square value.

The \emph{Bayesian Optimization Algorithm} (BOA)~\cite{Pelikan1999Boa} is a very general multivariate EDA and constructs an arbitrary dependency graph with respect to a metric of choice. If wanted, the degree of incoming edges, i.e., the number of dependent variables, can be limited. \citet{Pelikan1999Boa} propose the Bayesian Dirichlet metric as one possibility to determine the quality of a dependency graph, and they state that the general problem of finding an optimal graph is NP-hard. Thus, they suggest greedy algorithms or heuristics for efficiently creating good graphs.

%
%




%
%

\subsubsection{Other Classifications}
\label{sec:eda:otherClassifications}

Another approach to classify EDAs is not to differentiate them by how many dependencies they can model but by certain invariances that their probabilistic models may have.

One such classification stems from the theory of EAs and was introduced by \citet{LehreWittUnbiased2010}. The authors consider a new black-box complexity known as \emph{unbiased black-box complexity} in order to prove tighter lower bounds for commonly analyzed EAs. This definition is so general that it applies to any black-box algorithm optimizing pseudo-Boolean functions, thus, including EDAs.

Unbiased black-box complexity considers black-box algorithms optimizing perturbations of the hypercube, where a perturbation is any isometric automorphism of the hypercube.\!\footnote{Isometric automorphisms of the hypercube are all isomorphisms that permute any positions and may change a value of~$x$ into $1 - x$ per position.} For example, cyclicly shifting a bit string by one position to the right and changing the value of the first bit in the result is an isometric automorphism.

Given a fitness function and a perturbed variant of it, a black-box algorithm is said to be \emph{unbiased} if the queries to the black box in the perturbed setting are the same as the queries of the unperturbed setting when inverted with respect to the perturbation. Thus, an unbiased algorithm does not favor certain positions over other positions or~$1$s over~$0$s, or vice versa, i.e., it has no bias in this respect.

When considering general-purpose algorithms, unbiasedness is a nice property to have, as it certifies that the algorithm has no bias with respect to the encoding of the search space. However, when considering certain problems, different values may have a strict, different meaning, such that unbiasedness with respect to those values does not make sense.

All of the EDAs presented in Section~\ref{sec:eda:univariate} are unbiased when using uniform tie-breaking.
%

A seemingly similar but unrelated property that many EDAs feature is that their probabilistic model does not change, in expectation, if all samples have the same fitness, i.e., there is no signal from the fitness function. \citet{FriedrichKK16EDA} call this property \emph{balanced}, with respect to the \nblEDA. However, this property has been considered before already by \citet{ShapiroECJ05}, albeit with different terminology.

Although balancedness seems beneficial at first glance, it actually leads to the probabilistic model converging to one of the corners of the hypercube~\cite{ShapiroECJ05, FriedrichKK16EDA}. This is a general problem of martingales, i.e., random processes that do not change in expectation, with a bounded range, which will eventually end up at the bounds of their range. This means that balancedness implies a bias toward outer regions of the hypercube, also called \emph{genetic drift}~\cite{AsohMuehlenbeinDriftPPSN94}, as this is an inherent drift due to the genotypes of the sampled population. In order to overcome this bias and optimize successfully, the drift due to selection introduced by the fitness function has to be larger than the genetic drift.

Different approaches have been suggested in order to prevent an EDA's probabilistic model from  quickly converging to a corner of the hypercube. \citet{ShapiroECJ05} proposes to reject updates done to the probabilistic model with a probability equal to the ratio of going from one model to the other. This has the advantage that the resulting implicit distribution is the uniform distribution over the hypercube. However, the transition probabilities have to be known and computed in order to get the correct rejection probabilities. Another approach proposed by \citet{ShapiroECJ05} and also by \citet{FriedrichKK16EDA} is to introduce an artificial bias that counteracts the one introduced from the balancedness.

In the context of balancedness, \citet{FriedrichKK16EDA} introduce another concept, which they call \emph{stable}. An \nblEDA is stable if the limit distribution of each frequency, when no fitness signal is received, is unimodal with its maximum at~$1/2$. This means that a stable \nblEDA has a bias toward the center of the hypercube. The authors show that this concept is mutually exclusive with an \nblEDA being balanced, as those have a bias toward the corners of the hypercube. The \emph{stable} property is similar to the concept of an EDA's limit distribution being the uniform distribution, as considered by \citet{ShapiroECJ05}.

\subsection{Tools for Analyzing EDAs Theoretically}
Most of the theoretical results on EDAs consider univariate algorithms, as we explain in Section~\ref{sec:eda:runtime}. Thus, tools that make use of independent events are commonly used. However, that does not limit the use of these tools to the univariate case. Especially \emph{drift analysis}, which we present later in this section, can be applied in any setting.

Many proofs make use of classical probabilistic concentration bounds, such as Markov's inequality, Chebyshev's inequality, or, most importantly, Chernoff bounds~\cite{MitzenmacherUpfal2005}. The latter is used very frequently, since the sampling process of a univariate EDA is usually done independently of the other samples. Thus, the bound can be applied.

Since the theory on EDAs usually considers first hitting times, more specialized tools suited for that purpose are used, as well. One such tool is the Coupon Collector problem~\cite{MotwaniRaghavan1995}, which gives highly concentrated first hitting time results if a certain number of events with low probability have to occur to reach the target. For EDAs, this can be thought of as a certain number of factors of the probabilistic model being at the wrong end of their spectrum, thus, slowing down optimization, since they need to be changed for the optimization process to succeed.

Another tool for determining first hitting times, and the most prominent one when looking at theory on EAs and EDAs in general, is \emph{drift theory}. It is loosely akin to the potential method in complexity theory. To apply drift theory, one needs to define a potential that maps the stochastic process into the reals. Then, the expected difference of two consecutive steps of the process is considered: the \emph{drift}. This can be thought of as the expected velocity of the process. If the drift can be bounded, the expected hitting time of the process reaching a target is easily deducible, i.e., if there is a known bias in the process toward a certain direction, the first hitting time can easily be bounded.

We now state the three most commonly used drift theorems. The most general theorem with respect to the prerequisites of the process~-- the Additive Drift Theorem (Theorem~\ref{thm:eda:addDrift})~-- was stated by \citet{DBLP:journals/ai/HeY01}. However, the ideas used date back to Wald's equation~\cite{wald1944}.

\begin{theorem}[Additive Drift~\cite{DBLP:journals/ai/HeY01, DBLP:journals/nc/HeY04}]
    \label{thm:eda:addDrift}
    Let $(X_t)_{t \in \N}$ be random variables over a bounded space~$S \subseteq \R_{\geq 0}$ containing~$0$, and let $T = \min\{t \mid X_t = 0\}$.
    
    If there is a constant $\delta > 0$ such that, for all $t < T$,
    \[
        E[X_t - X_{t + 1}] \geq \delta\ , \textrm{ then } E[T \mid X_0] \leq \frac{X_0}{\delta}\ .
    \]
    And if there is a $\delta > 0$ such that, for all $t < T$,
    \[
        E[X_t - X_{t + 1}] \leq \delta\ , \textrm{ then } E[T \mid X_0] \geq \frac{X_0}{\delta}\ .
    \]
\end{theorem}

The Additive Drift Theorem can be applied when the expected difference of two potentials is known. However, oftentimes it is easier to determine the expected difference conditional on the current potential, i.e., $E[X_t - X_{t + 1} \mid X_t]$. Due to the law of total expectation, a lower bound for the conditional expected value is also a lower bound for the unconditional one.

A theorem more suited to processes whose potential changes at least linearly with respect to the current potential is the following Multiplicative Drift Theorem.

\begin{theorem}[Multiplicative Drift~\cite{DBLP:conf/gecco/DoerrJW10}]
    \label{thm:eda:multiDrift}
    Let $(X_t)_{T \in \N}$ be nonnegative random variables over~$\R$, each with finite expectation, and let $T = \min\{t \mid X_t < 1\}$.
    
    If there is a constant $\delta > 0$ such that, for all $t < T$,
    \[
        E[X_t - X_{t + 1} \mid X_t] \geq \delta X_t\ , \textrm{ then } E[T \mid X_0] \leq \frac{1 + \ln X_0}{\delta}\ .
    \]
\end{theorem}

The Multiplicative Drift Theorem is not well suited if the difference in potential is dependent on the current potential but not in a linear fashion. Such cases are covered by the following Variable Drift Theorem. However, note that all these theorems assume that the difference in potential does not increase when getting closer to the goal.

\begin{theorem}[Variable Drift~\cite{DBLP:journals/ijicc/MitavskiyRC09, Johannsen2010}]
    \label{thm:eda:variDrift}
    Let $(X_t)_{t \in \N}$ be nonnegative random variables over a bounded space~$S \subseteq \R_{\geq 0}$ containing~$1$, each with finite expectation, and let $T = \min\{t \mid X_t < 1\}$.
    
    If there exists a monotonically increasing function $h\colon \R_{\geq 0} \to \R_{\geq 0}$ such that $1/h$ is integrable and, for all $t < T$,
    \[
        E[X_t - X_{t + 1} \mid X_t] \geq h(X_t)\ , \textrm{ then } E[T \mid X_0] \leq \frac{1}{h(1)} + \int_{1}^{X_0}\frac{1}{h(x)}\,\mathrm{d}x\ .
    \]
\end{theorem}

The drift theorems above have been formulated in a simple, easy-to-read form that covers the 
most typical scenarios 
in which they are applied. However, more general drift theorems 
can be obtained \cite{LehreWittTailDriftArxiv,LehreWittISAAC2014}, \eg, to 
 apply Theorem~\ref{thm:eda:addDrift} in unbounded state spaces, Theorems~\ref{thm:eda:multiDrift} and~\ref{thm:eda:variDrift}
with respect to arbitrary minimum states $s_{\min}>0$ in the definition of~$T$ instead of state~$1$, and 
to allow processes adapted 
 to arbitrary stochastic filtrations instead of the natural one implicit in the formulations above.
These generalizations 
come partly 
at the expense of more complicated theorem statements, and sometimes require 
some additional technical assumptions on the underlying stochastic process.

\section{Common Fitness Functions}
\label{sec:eda:fitness_functions}

The most commonly analyzed pseudo-Boolean functions for EDAs are \om~\cite{Muehlenbein92} and \lo~\cite{RudolphDiss}. However, other functions have also been analyzed~\cite{ChenEtAlCEC09a, ChenEtAlCEC09b}, with \bv being the most prominent one of them~\cite{MuehlenbeinMahnigFDA-ECJ99, Droste2006a}.

\om counts the number of $1$s in a bit string. Thus, the unique optimum is the all-$1$s bit string.
\begin{equation}
    \label{eq:onemax}
    \om(\bm{x}) \coloneqq \sum_{i = 1}^{n} \bm{x}_i\ .
\end{equation}
This function can be generalized to a class of functions, each having a target bit string~$\bm{a}$~-- which denotes the unique global optimum~-- and yielding the number of incorrectly set bits.
Note that any unbiased algorithm, as introduced in Section~\ref{sec:eda:otherClassifications}, behaves on \om exactly as on the generalized version.

The \om function class is used to analyze how well an EDA performs as a hill climber. The usual expected runtime of an EDA on this function is $\Theta(n \log n)$~\cite{SudholtWittGECCO2016, KrejcaWittFOGA2017, WittGECCO17, LehreNguyenGECCO17}.

Whereas \om is oftentimes considered to be the easiest pseudo-Boolean function, \bv is said to be the hardest~\cite{Droste2006a}. In contrast to \om, where all bits are equally weighted, \bv uses exponentially scaled weights on its bit positions:
\begin{equation}
    \label{eq:binVal}
    \bv(\bm{x}) \coloneqq \sum_{i = 1}^{n} 2^{n - i}\bm{x}_i\ .
\end{equation}                                                  
That means that \bv represents value of a bit string interpreted as a binary unsigned integer.

Since the sum of all powers of~$2$ up to an exponent~$j$ is less than $2^j$\!, \bv can be interpreted as a lexicographic order on the hypercube, where lexicographically greater bit strings have a better fitness.

As for \om, in its general form, the global optimum of \bv is any bit string~$\bm{a}$, and the fitness of any bit string is the weight of the respective index if the bit value is the same as the one of~$\bm{a}$, and it is~$0$ otherwise.

\lo yields the number of consecutive $1$s in a bit string, starting from the left.
\begin{equation}
    \label{eq:leadingones}
    \lo(\bm{x}) \coloneqq \sum_{i = 1}^{n}\prod_{j = 1}^{i} \bm{x}_j\ .
\end{equation}
As in \om, the unique global optimum is the all-$1$s bit string. In its general version, the function yields the number of consecutively correctly chosen bits with respect to a fixed permutation~$\pi$ and a target bit string~$\bm{a}$.

\lo is used to analyze how an EDA copes with dependencies between the bits. The known expected runtime of certain EDAs on this function is $O(n^2)$ \cite{DangLehreGECCO15}, which is compliant with the usual upper bound of EAs on this function~\cite{AfshaniADDLM13}.

\section{Convergence Analyses}
\label{sec:eda:convergence}
The earliest theoretical studies of EDAs focused mostly on their convergence, similar in style to 
research that had been done for evolutionary algorithms in the 1990s \cite{RudolphDiss,VoseBook1999}. More precisely, 
it is studied how the algorithm behaves in the limit $t\to \infty$, \ie, if the algorithm is allowed to 
run for an arbitrary amount of time. If optimal solutions will be found 
in this limit, the algorithm is considered effective.

Almost all convergence analyses of EDAs consider univariate models. An early publication by \citet{HoehfeldRudolph1997PBIL} 
studies the vector of frequencies $\bm{p}^{(t)}$ in \pbil by a Markov-chain model 
and rigorously proves that if $\mu=1<\lambda$ and $\rho>0$, 
it in expectation will converge to a some solution $\bm{x}^*=(\bm{x}^*_1,\dots,\bm{x}^*_n)$, 
more precisely,  
 $E[\bm{p}_i^{(t)}]\to \bm{x}_i^*$ as $t\to \infty$. This solution need not be an optimal one but may correspond to a local optimum in which 
the search process is lead in the very first steps. If the fitness function $f$ is a linear pseudo-Boolean function,
 then in fact $E[\bm{p}_i^{(t)}]\to \bm{x}_i^*$ \wrt\ to 
the optimal solution $\bm{x}^*$\!. This includes classical benchmark functions like \om. However, as pointed out by 
\citet{ShapiroECJ05}, convergence in expectation does not imply that \pbil 
eventually will sample the optimum of such functions. In fact, genetic drift may lock frequencies to values 
that make it impossible to sample the optimum.

\pbil was also theoretically analyzed by \citet{Gonzales1999PBIL} using a dynamical systems model. Convergence of 
the model to local optima of the fitness functions is proven for $\mu=1$, and it is argued that the actual \pbil will 
resemble the model if $\rho$ is chosen sufficiently close to~$0$. Hence, the approach does not make predictions 
for high learning rates~$\rho$, in particular, it excludes the special case of $\rho=1$ as used in \umda.

Several subsequent works consider \umda and its generalization \fda. \citet{MuehlenbeinMahnigConvergenceFDA-jit-1999} also
use an approach similar to dynamical systems theory to derive a quantitative statement on the behavior of the frequencies in \fda and \umda   
over time. In fact, both fitness proportionate and the usual truncation selection (take the best $\mu$ out of $\lambda$ individuals) 
are considered. 
Specifically for the classical \umda on \om, they derive that roughly
\begin{equation}
\label{eq:eda:muehlenbein-formel}
\bm{p}_i^{(t+1)}\approx \bm{p}_i^{(t)} + \frac{I}{\sqrt{n}}\sqrt{\bm{p}_i^{(t)}\Big(1-\bm{p}_i^{(t)}\Big)}\ ,
\end{equation}
where $I$ is the so-called selection intensity that  is determined from the ratio $\mu/\lambda$ and can 
be thought of as being constant. By solving a differential equation, the formula can be turned into 
an approximation of the expected frequency at time~$t$. 
Interestingly, Equation~\eqref{eq:eda:muehlenbein-formel} resembles
 a rigorous statement on the drift of the frequencies that was recently 
proven in \cite{WittGECCO17} and is crucial for upper bounds on the runtime; see a more detailed discussion in 
Section~\ref{sec:eda:upper-onemax}.

A more comprehensive convergence study of the \fda is given 
by \citet{MuehlenbeinMahnigFDA-ECJ99}. 
As a general assumption, 
the \fda is instantiated with 
the correct decomposition of an additively decomposable function $f(x)=\sum_{i=1}^k f(X_j)$, where 
$X_j\subset\{1,\dots,n\}$, into its subfunctions. Then the algorithm will compute a probabilistic model, comprising 
unconditional and  conditional 
frequencies from the sampled search points. 
Strong results are obtained if a fitness-proportionate selection scheme called Boltzmann selection is used. 
Under some assumptions on the initial population, the algorithm will converge to a distribution that 
is uniform on the set of optimal solutions. The drawback of this result is that Boltzmann selection is 
computationally very expensive. For the usual truncation selection, results building on simplifying 
assumptions are obtained. Moreover, using infinite-population models,  
the paper derives quantitative 
statements similar to 
Equation~\eqref{eq:eda:muehlenbein-formel} on the time for a frequency of \umda to converge to its optimum value, 
regarding  
 \om and \bv.

In the 2000s, rigorous convergence proofs of the \fda (including \umda) with fitness-proportionate \cite{ZhangMIEEETEC2004} and 
truncation selection \cite{ZhangConvergenceFDA2004} followed. To study the regions of convergence and their stability,  
this research was supplemented by a fixed-point analysis for \umda and the \fda with 2-tournament selection 
in \cite{ZhangFixedPointsEDA2004}. It turns out that the \fda, given an appropriate decomposition 
of a non-linear function, converges under milder assumptions on the starting population than \umda.
Roughly, this indicates 
 that a multivariate model, as used in \fda, can be superior to a univariate model, to which 
\umda is restricted. However, 
also the analyses in \cite{ZhangMIEEETEC2004,ZhangFixedPointsEDA2004,ZhangConvergenceFDA2004} 
make the assumption of an infinite population size, which is very common in early convergence analyses 
of nature-inspired algorithms \cite{VoseBook1999}. Infinite populations simplify the analysis since 
certain stochastic effects leading to a deviation from the expected behavior, so-called fluctuations such as genetic drift, 
vanish under this assumption. 
Often this type of analysis 
is accompanied by experiments, which support the validity of the statements also for finite population sizes. Theoretically 
motivated research often demands rigorous statements that also hold for finite populations, see the following 
sections on runtime analysis.

%

%
%
%

%

A more recent work by Wu and Kolonko~\cite{WuKolonkoCrossEntropyConvergence2014}  
presents a convergence analysis of a so-called generalized cross-entropy optimization algorithm. 
The algorithm generalizes \pbil by adding so-called feasibility information to elements of the search space. This information corresponds 
to \emph{heuristic information} used in ACO~\cite{DorigoStuetzleACOBook}. 
It is shown for constant $\rho$ and under different assumptions on the feasibility 
information that the algorithm may stagnate in sub-optimal points due to genetic drift. 
However, for a time-dependent update scheme, almost sure convergence to a set of solutions that may include optimal points is proven. Finally, 
an initial runtime analysis on \lo is presented. However, this specific result 
is superseded by more detailed analyses in a follow-up paper \cite{WuKolonkoMoehringAnalysisCEIEEETEC} discussed below.

To conclude this overview of convergence analyses, we mention a very recent work by 
\citet{OllivierAAHIGO2017}. They introduce the \emph{Information-Geometric Optimization} algorithm, which is a very general EDA framework derived from three invariance properties: invariance under the parameterization of the search space, invariance under the parameterization of the probabilistic model, and invariance under monotone transformations of the fitness function. This means that IGO does not care about the encoding of the search space, the probabilistic model, or absolute fitness values. The authors show that IGO results in a general EDA that encompasses \pbil and \cga when it is used on the discrete hypercube considering Poisson binomial distributions. Further, they consider a time-continuous infinite-population version of IGO, which they call IGO flow, in the setting of linear pseudo-Boolean optimization and prove that it always converges to the optimum if the probabilistic model is not ill-initialized, i.e., none of the probabilities are initialized such that sampling the optimum is impossible.

\section{Runtime Analyses}
\label{sec:eda:runtime}
In contrast to convergence results as described in Section~\ref{sec:eda:convergence}, the focus of runtime analyses is the number of iterations until an algorithm samples a solution of sufficient quality for the first time, usually an optimum. Normally, the analyses consider both the expected number of iterations and concentration results.

In this section, we first give an in-depth overview about the history of runtime analyses on EDAs, ending with a very detailed discussion of the most recent results. These results are summarized in Table~\ref{tab:runtimes}. Then, we consider noisy scenarios, i.\,e., scenarios where the fitness function is perturbed by some kind of noise, usually as an additive term to the original fitness. In this setting, every time a solution is evaluated, the noise is drawn again and independently from any prior noise, and the goal is to optimize the underlying unperturbed function despite the noise.

\subsection{Early Results}
We start with a discussion of the first publications addressing runtime aspects of EDAs, which date back to the early 2000s. Although 
some of the runtime bounds proven in these publications can now be improved with state-of-the-art methods, the analyses already point out 
typical scenarios and challenges in the runtime behavior of EDAs, in particular regarding genetic drift. Also they give insights 
into fundamental properties of EDAs that distinguish them from other nature-inspired algorithms like EAs.


\subsubsection{First Steps Towards Runtime Analyses}

As pointed out above, rigorous runtime analyes must avoid the infinite-population model and derive 
statements for populations of finite sizes. However, the finiteness comes at a cost: if very small populations 
are used, there is a high risk of genetic drift and premature convergence in suboptimal regions 
of the search space. In a series of works, Shapiro \cite{ShapiroFOGA02,ShapiroECJ05,ShapiroPPSN06Diversity} 
addresses sources of 
 genetic drift in EDAs, quantifies its impact, and proposes measures to avoid it. 
 In \cite{ShapiroECJ05}, he points out that the probability distribution evolved by an EDA may converge to
suboptimal points and, using a dynamical systems approach, 
determines $\sqrt{n}$ as minimum population size for \umda to avoid genetic drift on the \om problem  and 
even exponential ones for \needle. 
Later, Sudholt and Witt~\cite{SudholtWittGECCO2016} and Krejca and Witt~\cite{KrejcaWittFOGA2017} gave 
rigorous proofs of the fact that genetic drift 
can happen up to population sizes of $O(\sqrt{n}\log n)$ in \cga and \umda. Alternatively, 
for \pbil, the learning rate~$\rho$ may be reduced  to counteract genetic drift. Using 
a dynamical systems approach, Shapiro \cite{ShapiroFOGA02} derives that the learning rate should be 
$O(1/\sqrt{n})$ and $O(2^{-n})$ to avoid genetic drift 
on the 
\om and \needle function, respectively.

In \cite{ShapiroPPSN06Diversity}, Shapiro also gives 
a rigorous theorem on the speed at which genetic drift moves the probabilistic model belonging  
to a specific class of EDAs called SML-EDA (including \umda) into suboptimal regions. 
Also, a rigorous bound $\Omega(2^{n/2}/\sqrt{n})$ is determined 
for the population size required to make genetic drift on \needle unlikely.


Finally, in \cite[p. 115]{ShapiroECJ05}, early conjectures on the runtime of \umda appear. 
More precisely, the paper experimentally determines a runtime of $\Theta(\lambda\sqrt{n})$ for \umda on \om (given that 
$\lambda$ is asymptotically larger than $\sqrt{n}$). This bound was  
 rigorously proven in \cite{WittGECCO17}. However, it should be noted that Shapiro's \umda slightly differs from the standard. 

\subsubsection{First Runtime Analyses}

The first rigorous runtime analysis of an EDA was given by \citet{Droste2006a}. He considers \cga without borders and proves the general 
lower bound $\Omega(K\sqrt{n})$ for its expected runtime on all linear functions. Using classical 
drift analysis and Chernoff bounds, Droste also proves 
the 
bound $O(K\sqrt{n})$ for \om, using $K=\Omega(n^{1/2+\epsilon})$, \ie, slightly 
above the threshold stated by \citet{ShapiroECJ05}. This bound becomes $O(n^{1+\epsilon})$ 
for the smallest $K$ covered by his analysis. 
 Finally, Droste argues that \textsc{BinVal} 
is more difficult to optimize than \om and asymptotically most difficult within the class 
of linear functions by proving that 
\cga without borders takes 
time $O(Kn)$ with at least constant probability on this function if $K=\Omega(n^{1+\epsilon})$,  and expected time 
at least $\Omega(Kn)$. The upper bound is $O(n^{2+\epsilon})$ for the smallest possible $K$ allowed. However, the 
lower bound $\Omega(Kn)$ does not come with a minimum value for $K$.



The results for \om were 
recently refined by \citet{SudholtWittGECCO2016}, using more advanced 
tools. In particular, all of Droste's upper bounds apply Chernoff bounds to show that genetic drift is unlikely, more 
precisely, he shows that the probability of a frequency dropping below~$1/3$ during the optimization is superpolynomially small.  Using 
a negative drift theorem, the upper bound is improved from $O(n^{1+\epsilon})$ to $O(n\log n)$ 
in \cite{SudholtWittGECCO2016}. 
See Section~\ref{sec:eda:upper-onemax} for more details. Regarding \textsc{BinVal}, a very recent 
analysis by \citet{WittGECCO18} proves Droste's conjecture that the function is harder to optimize than \om since 
the expected optimization time of \cga on \textsc{BinVal} is $\Omega(n^2)$ no matter how~$K$ is chosen. The 
idea of the analysis is to show for all $K=o(n)$ 
that genetic drift will lock many frequencies to~$0$ before the optimum 
can be found.

The results discussed in the previous two paragraphs are summarized in the following theorem.

\begin{theorem}[\cite{Droste2006a,WittGECCO18}]
    \label{thm:cgaBinVal}
Choosing $K=n^{1+\epsilon}$ for some constant $\epsilon>0$, 
the runtime of \cga without borders 
on \bv is bounded by $O(Kn)$ 
with probability $\Omega(1)$. Moreover, the expected runtime of \cga with and without borders 
on \bv is bounded from below 
by $\Omega(\min\{n^2,Kn\})$. 
\end{theorem}

In the years following Droste's seminal work, 
runtime analysis focused more on \umda and variants thereof. The first runtime analysis of a \umda variant was 
given by \citet{ChenEtAlCEC07}, who study the \lo function and a modification called \textsc{TrapLeadingOnes}. An 
expected optimization time of $O(\lambda n)$ of \umda
on \lo is derived under a the so-called \emph{no-random-error} assumption, which is similar to an infinite-population 
model and basically eliminates genetic drift. The 
authors also show that \textsc{TrapLeadingOnes}, which starts out in the same way as \lo but 
requires an almost complete change of the probabilistic model 
for the EDA to reach the global optimum,  
 yields expected exponential optimization time for \umda, using 2-tournament selection instead of the usual 
truncation selection. Moreover, a generalization of \umda 
similar to \pbil is considered, but it turns out that the strongest bounds apply for \umda.

Strictly speaking, \citet{ChenEtAlCEC07} only derived runtime bounds for a model of \umda. In subsequent work \cite{ChenEtAlCEC09b}, 
they therefore supplement a rigorous proof of the fact that \umda, when using appropriate borders for the frequencies, 
with high probability requires superpolynomial time to optimize \textsc{TrapLeadingOnes}. Similarly to Droste's early work, Chernoff bounds are 
applied to show that frequencies do not deviate much from their expected behavior, \ie, do not exhibit strong genetic drift. For 
the Chernoff bounds to be sufficiently strong, unusually large population sizes such as $\lambda=\Omega(n^{2+\epsilon})$ are required.

This approach was successfully picked up and extended 
in a more comprehensive journal publication \cite{ChenEtAlIEEETEC2010}. Using again $\lambda=\Omega(n^{2+\epsilon})$, 
the authors show that \umda without borders optimizes \lo in time $O(\lambda n)$ with overwhelming probability. Furthermore, the utility of 
appropriately set frequency borders is shown on a modification called \textsc{BVLO}, where the fitness landscapes requires the frequency of 
the last bit to be changed from one extremal value to the other one. Here \umda with borders has expected polynomial runtime, whereas 
\umda without borders with overwhelming probability will be stuck at non-optimal solutions.

Finally, using similar proof techniques, in particular Chernoff bounds,
\citet{ChenEtAlCEC09a} presented a constructed example function called \textsc{Substring}, 
on which simple EDAs and simple evolutionary algorithms behave fundamentally different. More precisely, 
it is proven that the \oea with any mutation probability $c/n$, where $c>0$ is constant, with 
overwhelming probability needs 
exponential time to find the optimum of the function, 
while \umda using $\lambda=\Omega(n^{2+\epsilon})$ and $\lambda/\mu=O(1)$ 
with very high probability finds the optimum in time $O(\lambda n)$. Specifically, it is beneficial for the optimization 
that \umda can sample search points with high variances as long as all frequencies are close to $1/2$. The \oea 
always samples with low variance in the vicinity of the best-so-far solution, which is detrimental on the specific 
example function.

\subsection{Recent Advances}
Only very few runtime analyses of EDAs were published in the years 2010--2014, most notably \cite{Neumann2010a,WuKolonkoCrossEntropyConvergence2014}. Starting from 2015, this research area gained  significant momentum again, 
see, \eg, \cite{DBLP:conf/isaac/FriedrichKKS15, DBLP:conf/gecco/FriedrichKKS15, DangLehreGECCO15}.
We now discuss the latest results in runtime analysis of EDAs. 
They mostly consider the standard benchmark function for EAs: \om. Using and advancing the toolbox for the analysis,  
matching upper and lower bounds are proven, giving a tight runtime result that allows a direct comparison 
of their performance with other nature-inspired algorithms.

\subsubsection{Upper Bounds for \om}
\label{sec:eda:upper-onemax}

Interestingly, early runtime analyses of EDAs focused more on variants of \lo instead of 
\onemax, which is the most commonly considered example function in evolutionary computation. In fact,  
the first runtime analysis of \umda on \om was not published until~2015 \cite{DangLehreGECCO15}. A possible 
explanation is that the hierarchical structure of \lo makes it more accessible to a runtime analysis than \onemax: if 
the best-so-far \lo-value is~$k$ and 
the frequencies of the first $k$ bits all have attained their maximum value, it is likely to sample 
only~$1$s there, which is typically needed for an improvement of the best seen function value. In contrast, 
there is no direct relationship between the \onemax-value and frequencies at specific bits. Also, modern 
runtime analyses of \umda \cite{DangLehreGECCO15, LehreNguyenGECCO17} reveal that a proof 
of runtime bounds for \lo can be relatively short and simple once the case of \om has been understood.

\paragraph{Results for \cga and \mmasib}
Before we describe the advances made in the runtime analysis of \umda in more detail, we 
discuss the state of the art for the simpler EDAs \mmasib and \cga. As mentioned above, Droste \cite{Droste2006a} 
showed that \cga typically optimizes \om in time $O(n^{1+\epsilon})$, using $K=n^{1/2+\epsilon}$\!. 
His variant of \cga does not use any borders on the frequencies, which is why he uses a comparatively 
large $K$ to make convergence of a frequency to~$0$ by genetic drift sufficiently unlikely. More recent analyses of \cga 
and also other EDAs like \umda 
mostly impose borders $\{1/n,1-1/n\}$ on the frequencies, as motivated in Section~\ref{sec:eda:univariate}.  
Using a more careful analysis of the stochastic behavior of 
frequencies, the classical $O(n\log n)$ runtime can be obtained, as shown in the following 
summary of theorems.

\begin{theorem}[\cite{Neumann2010a,SudholtWittGECCO2016}]
\label{thm:eda:upper-bound-cga-onemax}
If $\rho \le  1/(cn^{1/2}\log n)$ for a sufficiently large constant~$c > 0$ and $\rho \ge 1/\poly(n)$,
then \mmasib (with borders) optimizes \om in expected time~$O(\sqrt{n}/\rho)$. For $\rho = 1/(cn^{1/2}\log n)$, the runtime bound is~$O(n \log n)$.

The expected optimization time of \cga (with borders) on \om with $K\ge c\sqrt{n}\log n$ for a sufficiently
large $c>0$ and $K = \poly(n)$ is $O(\sqrt{n}K)$. This is $O(n\log n)$ for $K = c\sqrt{n}\log n$.
\end{theorem}

Theorem~\ref{thm:eda:upper-bound-cga-onemax} 
makes statements for two 
slightly different EDAs but the proofs of these statements follow roughly the same structure. Crucially, the effect 
of genetic drift is bounded: in the given time bound, \eg, $O(\sqrt{n}K)$ generations, the expected number 
of frequencies that drop below~$1/3$ is proven polynomially small, \eg, $O(1/n^2)$. Such a statement is typically 
obtained from a negative drift theorem. Next, the drift of frequencies towards~$1$ induced by 
selection (the so-called bias) is analyzed. It turns out that this bias is at least proportionate to the 
sampling variance of the EDA: roughly each frequency $\bm{p}_i$ increases by an expected amount of 
$O\big(\bm{p}_i(1-\bm{p}_i)/(K\sqrt{\sum_{j=1}^n \bm{p}_j})\big)$ in each generation. An analysis of this variable drift, using the variable 
drift theorem, then gives the desired runtime bound. (As variable drift analysis was not available to 
\citet{Neumann2010a}, a unified and simpler proof of the statement for \mmasib 
is given in \cite{SudholtWittGECCO2016}.)
In the unlikely event that a frequency has 
reached the wrong border~$1/n$ due to genetic drift, an event of probability $\Omega(1/n)$ is sufficient 
to lift the frequency again, which is absorbed in the total runtime due to the low expected number of such 
bad frequencies.

\paragraph{1st Phase Transition Around $\sqrt{n}\log n$}
Theorem~\ref{thm:eda:upper-bound-cga-onemax} requires $K\ge c\sqrt{n}\log n$. Recent research 
reveals that \cga in fact exhibits a phase transition in the regime $\Theta(\sqrt{n}\log n)$, similarly 
for \mmasib. 
If $K\le c'\sqrt{n}\log n$ for a sufficiently small constant~$c'>0$, then 
genetic drift will outweigh the drift due to selection such that a significant number of frequencies 
will drop to the lower border. In this case, classical arguments on coupon collector processes 
show that the runtime must be at least $\Omega(n\log n)$; see more arguments below in 
Section~\ref{sec:eda:onemax-lower} 
on lower bounds. There are no upper bounds on the runtime of \cga and \mmasib in the regime 
corresponding to $K\le c'\sqrt{n}\log n$, but it is conjectured that bounds resembling 
the existing ones for \umda (see Theorems~\ref{thm:eda:lehre-nguyen-upper} and \ref{thm:eda:witt-gecco17} below) 
can be obtained if $K\in [c_1\log n, c_2\sqrt{n}\log n]$ 
for appropriate constants $c_1,c_2>0$.

\paragraph{Results for \umda}
We complete this discussion of upper bounds by a review of recent advancements for \umda. As mentioned, 
\citet{DangLehreGECCO15} were the first to prove upper bounds for \umda on \om. If 
$\lambda\ge c\log n$ for a sufficiently large constant $c>0$ and $\lambda\ge 13\mathrm{e}\mu/(1-c')$ for 
an arbitrarily small constant $c'>0$, then the expected runtime of \umda on \om is $O(n\lambda \log \lambda)$. 
Hence, plugging in the smallest value of $\lambda$ allowed in the statement, the bound is $O(n\log n\log\log n)$, 
\ie, slightly above the $O(n\log n)$-bound discussed above \wrt\ \cga and \mmasib.

Dang and Lehre use a powerful proof technique to obtain their bound. 
Interestingly, the so-called level based theorem \cite{DangEtAlLevelBasedIEEETEC2017},
 which was originally developed for the analysis 
of population-based evolutionary algorithms, can be applied in this context. It is shown how the 
truncation selection of the best $\mu$ out of $\lambda$ individuals leads to a reasonable 
chance of improving the best-so-far \om value and allows to satisfy the other conditions of the 
level-based theorem under certain parameter settings. As a side-result, using the same proof technique, also the bound $O(n\lambda+n^2)$ \wrt the \lo function 
is obtained. 
Somewhat unusual, these proofs mostly 
consider populations instead of analyzing the values of 
single frequencies. For this to work, it is necessary that a frequency 
vector more or less unambiguously can be translated back into the population from which it was computed. 
This is possible in \umda but not even in the slight generalization \pbil, where a frequency vector depends 
on a history of previous populations.

Obviously, the proof of the above-mentioned $O(n\log n\log\log n)$ bound immediately raised the question whether 
this was the best possible runtime of \umda on \om. Recently, two independent improvements 
of the bound were presented. The first one due to Lehre and Nguyen \cite{LehreNguyenGECCO17} builds on 
a refinement of the level-based analysis,  
carefully using properties of the Poisson-binomial distribution, and is summarized by the following 
theorem. We emphasize that \umda always refers to the algorithm  with borders $1/n$ and $1-1/n$ on 
the frequencies, \ie,  
Algorithm~\ref{alg:eda:umda} extended by a step that narrows all frequencies down to the interval $[1/n,1-1/n]$.

\begin{theorem}[\cite{LehreNguyenGECCO17}]
\label{thm:eda:lehre-nguyen-upper}
For some constant $a > 0$ and any constant $c\in (0,1)$, \umda (with borders)
with parent population size $a\ln n\le \mu \le \sqrt{n(1-c)}$, 
offspring population size $\lambda \ge (13\mathrm{e})\mu/(1-c)$ has 
expected optimization time $O(n\lambda)$ on \om.
\end{theorem}

Hence, Theorem~\ref{thm:eda:lehre-nguyen-upper} proves 
that the runtime of \umda is $O(n\log n)$ for an appropriate choice of the parameters. This is tight 
due to the recent lower bound $\Omega(n\log n)$ discussed below in Section~\ref{sec:eda:onemax-lower}. 
Interestingly, the set of appropriate choices for the $O(n\log n)$ behavior
 is confined to $\lambda=\Theta(\log n)$, which 
corresponds to a parameter choice below the above-mentioned phase transition, \ie, a choice 
where the algorithm exhibits severe genetic drift. Also, the theorem includes a limit on $\mu$, 
which is exactly in the regime of the phase transition. For greater values 
of $\mu$ and $\lambda$, \citet{WittGECCO17} independently derived runtime bounds, see the following 
theorem; it also includes 
the regime covered by \citet{LehreNguyenGECCO17}, albeit with 
an assumption on the ratio $\lambda/\mu$.

\begin{theorem}[\cite{WittGECCO17}]
\label{thm:eda:witt-gecco17}
\begin{enumerate}
\item 
Let $\lambda=(1+\beta)\mu$ for an arbitrary constant~$\beta>0$ and let $\mu\ge c\sqrt{n}\log n$ for some sufficiently large 
constant~$c>0$. Then  the optimization time of \umda, both with and without borders, 
on $\om$ is bounded from above 
by $O(\lambda \sqrt{n})$ with probability $\Omega(1)$. For \umda with borders, also the expected optimization time is bounded in this way.

\item 
Let 
$\lambda=(1+\beta)\mu$ for an arbitrary constant~$\beta>0$ and $\mu\ge c\log n$ 
for a sufficiently large constant~$c>0$ as well as $\mu=o(n)$. Then the 
expected optimization time of \umda with borders on \om is $O(\lambda n)$.  
For \umda without borders, it is infinite with high probability 
if $\mu<c'\sqrt{n}\log n$ for a sufficiently small constant~$c'>0$.
\end{enumerate}
\end{theorem}

The two statements of Theorem~\ref{thm:eda:witt-gecco17} reflect the above-mentioned phase transition. 
For $\mu\ge c\sqrt{n}\log n$, as demanded in the first statement, the behavior is similar to the one underlying 
Theorem~\ref{thm:eda:upper-bound-cga-onemax} \wrt \cga and \mmasib. Frequencies move smoothly 
towards the upper border and it is unlikely that frequencies exhibit genetic drift towards 
smaller values than~$1/3$. Hence, it is unlikely as well that \umda without borders gets stuck 
with frequencies being at~$0$. The runtime $O(n\log n)$ is obtained for $\lambda=c\sqrt{n}\log n$ 
for an appropriately large constant~$c>0$.

The second statement of Theorem~\ref{thm:eda:witt-gecco17} applies to a case where genetic drift is likely, 
but frequencies that have hit the lower border~$1/n$ have a reasonable chance to recover in the given 
time span, which is $O(n\lambda)$ instead of only $O(\sqrt{n}\lambda)$ now. 
In fact, the analysis carefully considers the drift of frequencies from the lower towards the upper 
border and analyzes the probability that a frequency leaves its upper border again. To do so, a very careful 
analysis of the bias introduced by selecting the best $\mu$ individuals is required. Without such selection, 
a single frequency would correspond to a so-called martingale, 
but due to selection there is a small drift upwards, similarly 
to what we described \wrt \cga above. Hence, the proof of Theorem~\ref{thm:eda:witt-gecco17} 
also gives insights into the stochastic process 
described by single frequencies. It is more involved than the one for \cga since \umda can change 
frequencies globally instead of only by $\pm 1/K$. The runtime $O(n\log n)$ can be obtained again, this time  
for $\lambda=c\sqrt{n}\log n$.

It is worth pointing out that Theorems~\ref{thm:eda:lehre-nguyen-upper} 
and \ref{thm:eda:witt-gecco17} make non-overlapping statements. Theorem~\ref{thm:eda:witt-gecco17} also 
applies to $\lambda$ above the phase transition and describes a transition of $O(n\lambda)$ to $O(n\sqrt{\lambda})$ 
in the runtime. However, it crucially assumes $\lambda=(1+\Theta(1))\mu$ in both statements, an assumption that 
was also useful in earlier analyses of EDAs~\cite{ShapiroPPSN06Diversity} but restricts generality of the statements.
 In contrast, 
Theorem~\ref{thm:eda:lehre-nguyen-upper} 
applies to settings like $\mu=1$, $\lambda=c\log n$ and shows the $O(n\log n)$ bound also 
for this somewhat extreme choice of parameters.

We conclude this discussion of upper bounds by summarizing a recent study 
by \citet{WuKolonkoMoehringAnalysisCEIEEETEC}, who present the first runtime analysis of \pbil (called   
cross-entropy method (CE) in their paper). Using $\mu=n^{1+\epsilon}\log n$ for some constant~$\epsilon>0$ and 
$\lambda=\omega(\mu)$, they obtain that the runtime of \pbil on \om is $O(\lambda n^{1/2+\epsilon/3}/\rho)$ with overwhelming probability. 
Hence, if  $\rho=\Omega(1)$, including the special case $\rho=1$ where \pbil collapses to \umda, a runtime bound of $O(n^{3/2+(4/3)\epsilon}\log n)$ 
holds, \ie, slightly above $n^{3/2}$\!. In light of the detailed analyses of \umda presented above, one may conjecture that this bound is not tight 
even if $\rho<1$ is used, \ie, \pbil actually uses its learning approach to include solutions from several previous generations 
in the probabilistic model. In addition to that, a bound of the type $O(n^{2+\epsilon})$ on \lo is obtained if $\rho=\Omega(1)$, $\mu=n^{\epsilon/2}$ and 
$\lambda=\Omega(n^{1+\epsilon})$. Technically, \citet{WuKolonkoMoehringAnalysisCEIEEETEC} use concentration 
bounds such as Chernoff bounds to bound the effect of genetic drift 
as well as anti-concentration results, in particular for the Poisson-binomial distribution, to  obtain their statements. All bounds 
hold with high probability only since \pbil is formulated without borders. Probably, using a more detailed analysis of genetic 
drift and applying modern drift theorems, the bound for \lo can be improved to an expected $O(n^2)$ runtime 
for all $\rho=\Omega(1)$, provided 
that the classical borders $\{1/n,1-1/n\}$ are used.


\subsubsection{Lower Bounds for \om}
\label{sec:eda:onemax-lower}
Deriving lower bounds on the runtime of EDAs is often more challenging than deriving upper bounds. Roughly, most existing approaches 
show that the probabilistic model is not sufficiently adjusted towards the set of optimal solutions within a given time span. A relatively 
straightforward approach relates the runtime to the strength of updates of the algorithm. With respect to simple univariate algorithms like \cga and \umda, one can show that frequencies do not increase more than 
by $1/K$ (with probability~$1$) resp.\ $O(1/\mu)$ (in expectation, assuming $\lambda=(1+\Theta(1))\mu$) in 
a step. This naturally leads to a lower bound 
of $\Omega(K)$ resp.\ $\Omega(\mu)$ on the runtime on \om. However, the bound is weak as it pessimistically assumes 
that each generation changes frequencies in the right direction. More detailed analyses reveal that \cga 
in the early phases of the optimization process only has 
a probability of $O(1/\sqrt{n})$ of performing a step where the two offspring 
differ in less than two bits, \ie, the probability 
that the outcome of a certain bit is relevant for  
selection is then only $O(1/\sqrt{n})$ \cite{SudholtWittGECCO2016}. Similar results can be 
obtained \wrt\ \umda \cite{KrejcaWittFOGA2017}. 
Thus, each bit only moves up to an expected amount of $O(1/(K\sqrt{n}))$ resp.\ $O(1/(\mu\sqrt{n}))$ per generation. 
Then a drift analysis translates this into the lower bounds $\Omega(K\sqrt{n})$ resp.\ $\Omega(\mu\sqrt{n})$ that 
appear in the following theorems. The first bound was already known for \cga without borders from Droste's work \cite{Droste2006a}.

\begin{theorem}[cf.~\cite{SudholtWittGECCO2016}]
\label{theo:eda:lower-cga-om}
The optimization time of \cga (with borders) with $K \le \poly(n)$ on \om 
is $\Omega(K\sqrt{n} + n \log n)$ with high probability  and in expectation.
\end{theorem}

\begin{theorem}[\cite{KrejcaWittFOGA2017}]
\label{theo:eda:lower-umda-om}
Let $\lambda=(1+\beta)\mu$ for some constant $\beta>0$ and $\lambda\le \poly(n)$. 
Then the expected optimization time of \umda on $\om$  is $\Omega(\mu \sqrt{n} + n \log n)$ (both with and without borders). 
\end{theorem}

Sudholt and Witt 
\cite{SudholtWittGECCO2016} also state Theorem~\ref{theo:eda:lower-cga-om} 
in an analogous fashion for 
\mmasib, with the parameter $K$ replaced by $1/\rho$. As its working principle is rather 
similar to \cga, we do not discuss 
\mmasib further in this section.

The lower bounds $\Omega(K\sqrt{n})$ and $\Omega(\mu\sqrt{n})$ we illustrated so far are very weak 
if $K$ resp.\ $\mu$ are small. In fact, they can be even worse than the bounds $\Omega(n/\!\log n)$ 
that follow from black-box complexity \cite{DJWBlackBox}. Until 2016, it was not clear whether 
the runtime of these simple EDAs also was bounded by $\Omega(n\log n)$ or whether they could possibly 
optimize \om in $o(n\log n)$ time and hence be faster than simple evolutionary algorithms. 
A negative answer was given by the two above theorems, both of which 
also contain an $\Omega(n\log n)$ term. 

The proof of the bound $\Omega(n\log n)$ is technically demanding. It relies on the following strategy:
\begin{enumerate}
\item 
\label{it:eda:lower-bound-om-freq-down}
Show that with high probability several frequencies, \eg, $\sqrt{n}$ many, reach the lower border before the optimum is 
sampled. This requires a detailed analysis of the stochastic behavior of several dependent, single frequencies 
instead of considering merely the sum $P_t\coloneqq \sum_{i=1}^n \bm{p}^{(t)}_{i}$ of 
the frequencies, whose stochastic behavior is already quite well understood and can relatively easily be 
analyzed by drift analysis, as sketched in the paragraph following Theorem~\ref{thm:eda:upper-bound-cga-onemax}.
 In fact, in the detailed analysis of single frequencies, it is even required to show that 
some frequencies walk to the lower border while most other frequencies do not move up too far 
to the upper border; otherwise 
one cannot rule out with sufficiently high probability that the optimum is sampled in the meantime.
\item
Once polynomially many frequencies have reached the lower border~$1/n$, a so-called coupon collector effect arises. 
A relatively straighforward generalization of the coupon collector theorem \cite{MotwaniRaghavan1995,MitzenmacherUpfal2005} to the case 
that still polynomially many bits have to be corrected, where a correction 
is made with probability at most $1/n$, yields the following statement: \emph{
Assume \cga reaches a situation where at least $\Omega(n^\epsilon)$ frequencies attain the lower border~$1/n$. Then with high probability and in expectation, the remaining optimization time is $\Omega(n \log n)$.}
The underlying modification of the coupon collector theorem may be called folklore in 
probability theory, but interesting for its own sake: collecting the last $n^\epsilon$ coupons 
takes asymptotically the same time as collecting them all.
\end{enumerate}

A major effort is required to flesh out the behavior sketched in Item~\ref{it:eda:lower-bound-om-freq-down} above. Roughly 
speaking, it is exploited that frequencies behave similarly to a martingale and can walk to the lower border due to genetic drift. 
However, the effect of genetic drift is dependent on many factors. When all frequencies have reached a border, genetic 
drift is much less pronounced than in situations where many frequencies are close to the medium value~$1/2$ (which is initially 
the case). To handle this dependency on time, it is shown that some frequencies move unusually fast, which 
means faster than the expected time, to the lower border 
while the majority of the frequencies is still at a medium value. More precisely, the proofs approximate the hitting time 
of the lower border by a normally distributed random variable, which is not sharply concentrated around this mean 
and exhibits exactly the desired reasonable probability of deviating from the mean. Additionally, the drift analysis 
features a novel use of potential functions that smooth out the variances of the movements of frequencies, which would be 
place-dependent and not applicable to the approximation by a normal distribution otherwise.

\paragraph{2nd Phase Transition Around $\log n$}
Not much research has been done on very small values of the population size $\lambda$ and $K$ in 
\umda and \cga, corresponding to very big $\rho$ in \mmasib. \citet{Neumann2010a} 
give an exponential bound on the runtime of \mmasib if $\rho\ge c/\log n$, indicating a second 
phase transition in behavior around $\log n$. Roughly speaking, if the set of possible 
values for a frequency becomes less than $\log n$, then the scale is too coarse for 
the probabilistic model to adjust slowly towards the set of optimal solutions. For example, 
even after a frequency has reached its maximum $1-1/n$ once, an unlucky step may lead to 
a drastic decline in frequency which on average cannot be recovered in polynomial time. It is 
conjectured that \cga and \umda will not optimize \onemax in polynomial time either 
if $K\le c\log n$ resp.\ $\lambda\le c\log n$ for a small constant~$c>0$.

\paragraph{Major Open Problems} 
Even if we ignore the values below $\log n$ corresponding to the second phase transition 
just mentioned, 
the lower bounds given in Theorem~\ref{theo:eda:lower-cga-om} and \ref{theo:eda:lower-umda-om} still do not 
give a complete picture of the runtime of the algorithms on \om. For example, for $\mu$ being in the medium 
regime between the phase transitions, \ie, $\mu$ being both $\omega(\log n)$  
and $o(\sqrt{n}\log n)$, it is not clear whether a lower bound of the  kind $\Omega(\mu n)$ (which 
would match the upper bound given above in Theorem~\ref{thm:eda:witt-gecco17}) or any other runtime being 
$\omega(n\log n)$ holds. It is an open problem to prove tight bounds on the runtime of  
the simple EDAs in this medium regime. 
As usual, we expect analyses to be harder for \umda than for \cga, as the former algorithm 
can change frequencies in a global way, while the latter only changes locally by $\pm 1/K$.

Some progress on the way to tight bounds 
has been made very recently by Lengler et al.~\cite{LenglerSudholtWittGECCO18}, who prove a lower 
bound of $\Omega(K^{1/3}n+n\log n)$ on the expected optimization time of  
the \cga if $K=O(n^{1/2}/(\log n\log\log n))$. Hence, the expected optimization time will be $\Omega(n^{7/6}/(\log n\log\log n))$ for $K=O(n^{1/2}/(\log n\log\log n))$, while it is bounded from above by $O(n\log n)$ for $K=cn^{1/2}$ if 
$c$ is chosen as
 a sufficiently large constant. Hence, the runtime seems to depend in a multimodal way on~$K$. Nevertheless, 
this still remains a conjecture since there are no upper bounds on the runtime of the \cga for $K=o(n^{1/2})$; there 
are only upper bounds for the \umda if $\lambda=o(n^{1/2})$ that support this conjecture.

\begin{figure}
\centering\begin{tikzpicture}[yscale=0.70,xscale=1.5]

\draw[->] (0,0) -- (6,0) node[anchor=north, above=0mm, right=1mm] {$\lambda$};
\draw	(0,0) node[anchor=north] {0}
     (1,0) node[anchor=north] {$\log n$}
		(3,0) node[anchor=north] {$\sqrt{n}\log n$};

\draw[->] (0,0) -- (0,4.5) node[anchor=east] {\scriptsize  runtime};

\node [rotate=90] at (0.5,2.2) {\color{red}{exponential} {\tiny (conjectured)}};
\draw[dotted, thick] (2.95,0) -- (2.95,4.5);
 \draw[dotted, thick] (3.05,4.5) -- (3.05,0);
\draw[dotted, thick] (0.95,0) -- (0.95,4.5);
 \draw[dotted, thick] (1.05,4.5) -- (1.05,0);

\draw[color=black!20,fill] (0.95,0.02) -- (0.95,4.5) -- (1.05,4.5) -- (1.05,0.02);

\draw[color=black!20,fill] (2.95,0.02) -- (2.95,4.5) -- (3.05,4.5) -- (3.05,0.02);

\begin{scope}
\clip (0,0) rectangle (0.95,5);
\draw[thick,dotted] (0,4.5) edge [bend right] (2.95,3.3);
\end{scope}

\draw [thick] (1.05,1) -- +(1.9,2) node[pos=0.5,above,rotate=25]{$O(\lambda n)$} node[pos=0.5,below,rotate=25] {\tiny (proven)};

\draw[thick] (3.05,1) -- +(2.95,1.5) node[pos=0.5,above,rotate=13]{$O(\lambda \sqrt{n})$} node[pos=0.5,below,rotate=13] {\tiny (proven)};

\end{tikzpicture}
\caption{Picture of runtime bounds for \umda on \om, assuming $\lambda=(1+\Theta(1))\mu$.}
\label{fig:eda:landscape-runtime-umda-onemax}
\end{figure}
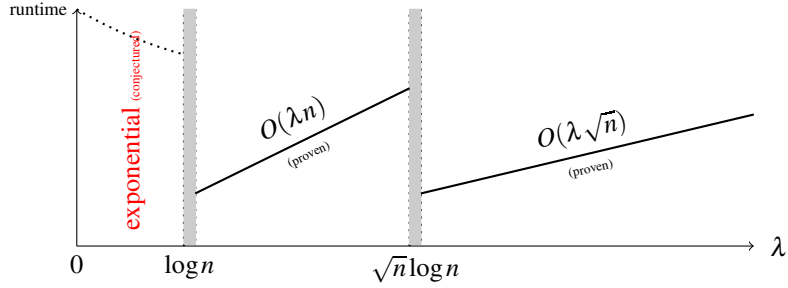

A summary of proven upper and conjectured bounds for the runtime of \umda on \om is displayed in 
Figure~\ref{fig:eda:landscape-runtime-umda-onemax}. We believe that similar results hold 
for \cga and \mmasib, with $\lambda$ replaced by $K$ and $1/\rho$, respectively.

We have carried out experiments for \umda on \om to gain some empirical insights into the relationship 
between $\lambda$ and the runtime. The algorithm was implemented in the C programming language 
using the WELL512a random number generator. 
The problem size was set to $n=2000$, $\lambda$ was increased from $14$ to $350$ in steps of size~$2$, $\mu$ 
was set to $\lambda/2$, and, due to the high variance of the runs especially for small $\lambda$, 
an average was taken over $3000$ runs for every setting of~$\lambda$. The left-hand side 
of Figure~\ref{fig:eda:experiments} demonstrates that the runtime in fact shows a multimodal dependency on~$\lambda$. 
Starting out from very high values, it takes a minimum at $\lambda\approx 20$ and then increases again up to $\lambda\approx 70$. Thereafter 
 it falls 
again up to $\lambda\approx 280$ and finally increases rather steeply for the rest of the range. The right-hand side also illustrates that 
the number of times the lower border is hit seems to decrease exponentially with $\lambda$. The phase transition where the behavior 
of frequencies turns from chaotic into stable is empirically located somewhere between $250$ and $300$.

\subsubsection{New Advances in Tackling Genetic Drift}
Genetic drift slows down optimization because it basically adds a random signal to the objective function. One reason this impacts the algorithms is their myopic behavior: they have to perform an update to their frequencies based only on information from the current iteration. Especially if this sample size is small, like for \cga or \mmasib, the information gained during a single iteration may be too small to perform a sensible decision with respect to the update.

In order to counteract such ill-informed updates, \citet{DoerrKrejcaGECCO18} propose a new EDA that tries to reduce the number of incorrect frequency updates by not only relying on the information from a single iteration but from multiple previous iterations. Their \emph{significance-based cGA} (\sigcGA) stores a frequency vector, like an \nblEDA, but additionally also stores a history~$H_i$ for each bit position~$i$. Each iteration, only two offspring are sampled, and the bits of the better individual are saved in the respective histories. Then the algorithm checks for each history whether a \emph{significance} occurs, that is, whether the number of $1$s or $0$s saved is drastically more than expected when assuming that each $1$ occurs with probability $\bm{p}_i$. The level of confidence can be regulated by a parameter called~$\varepsilon$. If a significance of $1$s is detected at a position, the respective frequency is set to $1 - 1/n$, if a significance of $0$s is detected, the frequency is set to $1/n$; otherwise it is left unchanged. Overall, the algorithm only uses three different frequency values: $1 - 1/n$, $1/2$, and $1/n$, where $1/2$ is only used as a starting value~-- if a frequency once takes a value different from $1/2$, it never returns.

This significance-based approach allows \sigcGA at the beginning of an optimization to keep frequencies at~$1/2$ until there is statistical proof that another value would be more beneficial. Thus, it can be thought of as an algorithm that is both balanced and stable.\!\footnote{Since \sigcGA is not an \nblEDA (due to the histories that store data from multiple iterations), the actual definitions of \emph{balanced} and \emph{stable} do not apply.} The usefulness of this approach was shown by proving that this algorithm optimizes \om and \lo both in time $O(n\log n)$ in expectation and with high probability, which has not been proven for any other EA or EDA before~\cite{DoerrKrejcaGECCO18}.

\pgfplotstableread[col sep = semicolon]{umda2000-3000.csv}\mydata
\begin{figure}
\centerline{
\begin{tikzpicture}[scale=0.7]
  \begin{axis}[
    legend pos = north east, 
    xmin = 14, xmax = 350, xlabel = $\lambda$
    ]
    \addplot table[x index = {0}, y index = {1}, mark = none]{\mydata};
		\end{axis}
		\begin{scope}[xshift=8 cm]
		\begin{axis}[
    legend pos = north east,
    xmin = 25, xmax = 350, xlabel = $\lambda$
    ]
		    \addplot table[x index = {0}, y index = {2}, mark = none]{\mydata};
  \end{axis}
	\end{scope}
\end{tikzpicture}}

\caption{Left-hand side: empirical runtime of \umda on \om, right-hand side: number of hits of lower border; for $n=2000$, $\lambda\in\{14,16,\dots,350\}$, $\mu=\lambda/2$,  and averaged over 
$3000$ runs}
\label{fig:eda:experiments}
\end{figure}
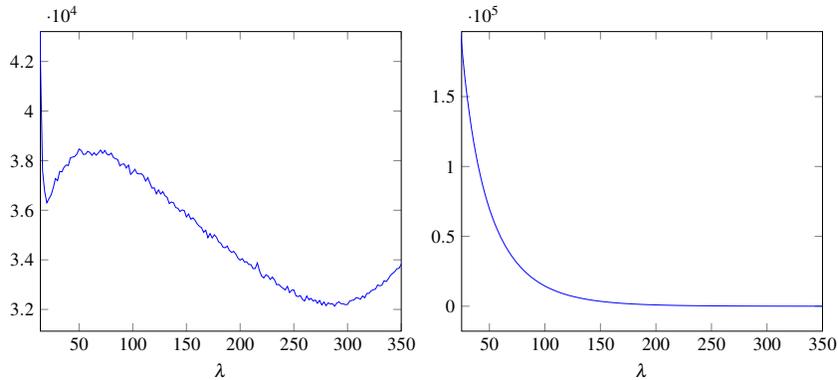

\begin{landscape}
    \pagestyle{empty}
    \begin{adjustwidth*}{-1.5 cm}{0 cm}
        \begin{table*}
            \caption{Expected run times (number of fitness evaluations) of various EDAs until they first find an optimum for the three functions \om (eq.~\eqref{eq:onemax}), \lo (eq.~\eqref{eq:leadingones}), and \bv (eq.~\eqref{eq:binVal}).
            }
            \hspace*{-0.2 cm}\begin{tabular}{lp{3.8 cm}p{4.4 cm}p{3.2 cm}p{2 cm}p{2.4 cm}p{2 cm}}\midrule
                \label{tab:runTimeComparison}
                Algorithm & \om & constraints & \lo & constraints & \bv & constraints\\ \toprule
                %
                %
                %
                %
                
                
                UMDA/PBIL\footnote{The results shown for PBIL are the results of UMDA, since the latter is a special case of the former. \citet{WuKolonkoMoehringAnalysisCEIEEETEC} also analyze PBIL but with worse results.} & $\Omega(\lambda\sqrt{n} + n\log n)$ (Thm.~\ref{theo:eda:lower-umda-om}) & $\mu = \Theta(\lambda), \lambda = O\big(\poly(n)\big)$ & $O(n\lambda\log\lambda + n^2)$~\cite{DangLehreGECCO15} & $\lambda = \Omega(\log n),\newline \mu = \Theta(\lambda)$ & unknown & --\\
                
                & $O(\lambda n)$ (Thm.~\ref{thm:eda:lehre-nguyen-upper} and~\ref{thm:eda:witt-gecco17}) & $\mu = \Omega(\log n) \cap O(\sqrt{n}), \lambda = \Omega(\mu)$ or $\mu = \Omega(\log n) \cap o(n), \mu = \Theta(\lambda)$ & & & &\\
                
                & $O(\lambda\sqrt{n})$ (Thm.~\ref{thm:eda:witt-gecco17}) & $\mu = \Omega(\sqrt{n}\log n), \mu = \Theta(\lambda)$ & & & &\\
                
                cGA/2-MMAS$_\textrm{ib}$ & $\Omega\Big(\frac{\sqrt{n}}{\rho} + n\log n\Big)$ (Thm.~\ref{theo:eda:lower-cga-om}) & $\frac{1}{\rho} = O\big(\mathrm{poly}(n)\big)$ & unknown & -- & $\Omega(\min\{n^2, Kn\})$ (Thm.~\ref{thm:cgaBinVal})\footnote{\label{fn:cgaBinVal}This result was only proven for cGA.} & none\\
                
                & $O\Big(\frac{\sqrt{n}}{\rho}\Big)$ (Thm.~\ref{thm:eda:upper-bound-cga-onemax}) & $\frac{1}{\rho} = \Omega(\sqrt{n}\log n) \cap O\big(\mathrm{poly}(n)\big)$ & & & $O(Kn)$ (Thm.~\ref{thm:cgaBinVal})\textsuperscript{\ref{fn:cgaBinVal}} & $K = n^{1 + \varepsilon},\newline \varepsilon = \Theta(1)$\\
                
                %
                
                
                \sigcGA & $O(n\log n)$ \cite{DoerrKrejcaGECCO18} & $\varepsilon > 12$ & $O(n\log n)$ \cite{DoerrKrejcaGECCO18} & $\varepsilon > 12$ & unknown & --\\ \midrule
            \end{tabular}
            \label{tab:runtimes}
        \end{table*}
    \end{adjustwidth*}
\end{landscape}

\subsection{Noisy Settings}
In real-world optimization, the evaluation of a solution often involves a degree of uncertainty due to inaccuracies in the evaluation process. We call this uncertainty in the fitness \emph{noise}. Since EDAs as general-purpose heuristics build on these inaccurate information, it is interesting to analyze how they perform when facing noise.

Most EDA scenarios with noise consider ACO variants on single-destination shortest-path problems, mostly not in the context of EDAs at all \cite{Horoba:2010:ACO:1830483.1830750, SudholtThyssen2012, Feldmann:2013:OEP:2460239.2460246, Doerr:2012:AES:2330163.2330167}. However, some results analyze pseudo-Boolean optimization~\cite{DBLP:journals/ec/FriedrichKKS16, friedrichKKS2017compact}.

\subsubsection{Combinatorial Optimization}
\citet{Horoba:2010:ACO:1830483.1830750} consider an acyclic weighted graph and are interested in finding a shortest path from each vertex to a single given destination. The noise is modeled by drawing a random nonnegative value~$\eta$ per edge weight~$w$, possibly dependent on the edge, and determine its new weight~$w'$ by $w' = w(1 + \eta)$. Thus, depending on the distribution of~$\eta$, large weights increase more than small weights. The algorithm of interest is an ACO variant. It constructs paths from each node to the destination, using the perturbed weight and choosing an edge with a probability relative to its pheromone value with respect to the pheromones of all competing edges. The algorithm compares each constructed path with the currently best-so-far solution per node without re-evaluation. That means that the best-so-far solutions as well as their possibly perturbed weights are stored and used for look-up.

The authors provide instances on which the algorithm does not find a desired approximation within polynomial time with high probability. This is due to the best-so-far solution not being re-evaluated. Thus, if a non-optimal path is evaluated to be very good by chance, it will get reinforced many times, making it more unlikely to sample other paths that, additionally, have to be evaluated even better. However, the authors also prove that optimization will succeed if the noise follows the same distribution for every edge.

\citet{SudholtThyssen2012} extend the results from \citet{Horoba:2010:ACO:1830483.1830750} by considering a vaster range of noise distributions, showing how long it takes to approximate optimal solutions or even when optimization succeeds.

\citet{Doerr:2012:AES:2330163.2330167} consider a similar scenario to the one analyzed by \citet{Horoba:2010:ACO:1830483.1830750}, the difference being that the weights of the graphs are purely random, i.e., there is no ground-truth to rely on. This setting makes it harder to define what an optimal solution actually is.

The authors first consider a multigraph consisting of two nodes with multiple edges between those nodes. They call an edge \emph{preferred} if its probability of being shorter than any other edge from the same vertex is at least $1/2 + \delta$, where $\delta > 0$ is a constant, and they state how this scenario relates to armed-bandit settings. Using the same ACO algorithm as \citet{Horoba:2010:ACO:1830483.1830750} but re-evaluating the best-so-far solution each iteration, the authors give an upper bound on the expected time until the pheromone on the preferred edge is maximal. They then provide examples of weight distributions that result in an edge being preferred. The paper concludes with a more general graph setting that assumes that there exists an inductively defined set of edges~$S$, starting at a given node, such that each edge extending paths using edges from~$S$ is preferred. If~$S$ is a tree, the authors give an upper bound for the expected time until the considered ACO variant maximizes the pheromones on all of the edges in~$S$.

\citet{Feldmann:2013:OEP:2460239.2460246} analyze the same setting as \citet{Doerr:2012:AES:2330163.2330167} but investigate another ACO variant: MMAS-fp. This algorithm does not store best-so-far solutions but always makes an update with respect to the current samples; however, the update is done with respect to each sample's fitness. Thus, good solutions yield larger changes in the update than bad solutions. The authors explain the difference in this approach with respect to the works of \citet{Horoba:2010:ACO:1830483.1830750} and \citet{Doerr:2012:AES:2330163.2330167} that MMAS-fp optimizes paths that are shortest in expectation. They prove this claim by providing upper bounds on the expected number of iterations until MMAS-fp finds expected shortest paths in graphs where, for each node, the difference between the expected lengths of different outgoing edges can be lower-bounded by a value $\delta > 0$, which influences the runtime.

\subsubsection{Pseudo-Boolean Optimization}
\citet{DBLP:journals/ec/FriedrichKKS16} (conference version~\cite{DBLP:conf/gecco/FriedrichKKS15}) also consider MMAS-fp, just like \citet{Feldmann:2013:OEP:2460239.2460246}, but in the setting of optimizing linear pseudo-Boolean functions. The noise is mostly modeled as Gaussian \emph{additive posterior noise}, i.e., when evaluating the fitness of an individual, a normally distributed random variable is added to the fitness, every time anew and independently. The authors show that MMAS-fp does, what they call, \emph{scale gracefully} in this scenario. That means that for every polynomially bounded variance of the noise, there is a configuration of MMAS-fp such that the runtime is polynomially bounded as well. Since the runtime results hold with high probability, by performing an uninformed binary search using restarts, the correct variance of a problem with Gaussian noise can be guessed correctly within polynomial time. Thus, MMAS-fp can be modified such that the runtime is, with high probability, polynomial if the variance of the noise is.

Additionally, the authors extend their results to other posterior noise than Gaussian. Further, they consider a \emph{prior noise} model where, before evaluating the fitness of an individual, a uniformly randomly chosen bit is flipped. In both of these settings, they prove that the algorithm scales gracefully.

\citet{friedrichKKS2017compact} (conference version~\cite{DBLP:conf/isaac/FriedrichKKS15}) also consider \cga under the Gaussian additive posterior noise model. As for MMAS-fp, they prove that the algorithm scales gracefully. Further, they show that the $(\mu + 1)$~EA, a commonly analyzed EA, does not scale gracefully. Both results of \citet{DBLP:journals/ec/FriedrichKKS16, friedrichKKS2017compact} suggest that EDAs are inherently more tolerant to noise than standard EAs, as the EDAs did not need to be modified to cope with noise, except for choosing correct parameters. The authors also compare the restart version of \cga with an approach that uses resampling in order to basically remove the noise in the fitness, as described by~\citet{akimoto:hal-01194556}. Since the number of resamples is closely tied to the noise's variance, the \cga variant using restarts instead of resampling emerges victorious.

\section{Conclusions and Open Problems}
\label{sec:eda:conclusions}

We have given an overview of the state of the art in the theory of discrete EDAs, where 
the most recent research surpasses convergence analyses and instead 
deals with the runtime 
of especially simple univariate EDAs like \cga, \umda, and \pbil. In this domain, 
increasingly precise results have been obtained with respect to well-established benchmark problems 
like \onemax, but as we have emphasized in this article, there are several open problems 
even for this simple problem. In particular, a complete picture of the runtime 
of the simple EDAs depending on their parameters is still missing. We think that 
further results for benchmark functions will give insight into the right choice 
of specific EDAs, including the choice 
of parameters such as the population size and the borders on the frequencies depending 
on problem characteristics. We also expect that this research will lead to runtime results 
and advice on the choice of algorithms and parameters \wrt 
more practically relevant combinatorial optimization problems. Here in particular, 
noisy settings or, more generally, optimization under uncertainty 
seem to represent scenarios where EDAs can outperform classical evolutionary 
algorithms. Also, the combinatorial structure may favor the application of multivariate 
EDAs, a type of EDAs for which almost no theoretical results exist yet.

%
%
%
%
%
%
%

\section*{Acknowledgments}
Carsten Witt was supported by a grant by the Danish Council for Independent Research (DFF-FNU 4002-00542). Support by the COST Action 15140 ``Improving Applicability of Nature-Inspired Optimisation by Joining Theory and Practice (ImAppNIO)'' is also gratefully acknowledged.
%
%
%
\bibliographystyle{spbasic}
\bibliography{eda}
\end{document}